\pdfoutput=1

\documentclass[11pt]{article}

\usepackage{EMNLP2022}

\usepackage{times}
\usepackage{latexsym}

\usepackage[T1]{fontenc}

\usepackage[utf8]{inputenc}

\usepackage{microtype}

\usepackage{inconsolata}

\usepackage{graphicx}
\usepackage{times}
\usepackage{epsfig}
\usepackage{amsmath}
\usepackage{amssymb}
\usepackage{threeparttable}
\usepackage{lscape}
\usepackage{array}
\usepackage{booktabs}
\usepackage{caption}
\usepackage{subcaption}
\usepackage{bm}
\usepackage{multirow}

\newcommand{\method}{\textit{DiffG-RL}}

\title{\method: Leveraging Difference between State and Common Sense}

\author{Tsunehiko Tanaka \thanks{\: Work done during an internship at IBM Research.} \\
  Waseda University \\
  \texttt{tsunehiko@fuji.waseda.jp} \\\And
  Daiki Kimura \\
  IBM Research \\
  \texttt{daiki@jp.ibm.com} \\\And
  Michiaki Tatsubori \\
  IBM Research \\
  \texttt{mich@jp.ibm.com}}

\begin{document}
\maketitle
\begin{abstract}
Taking into account background knowledge as the context has always been an important part of solving tasks that involve natural language. One representative example of such tasks is text-based games, where players need to make decisions based on both description text previously shown in the game, and their own background knowledge about the language and common sense. In this work, we investigate not simply giving common sense, as can be seen in prior research, but also its effective usage. We assume that a part of the environment states different from common sense should constitute one of the grounds for action selection. We propose a novel agent, \method, which constructs a Difference Graph that organizes the environment states and common sense by means of interactive objects with a dedicated graph encoder. \method~also contains a framework for extracting the appropriate amount and representation of common sense from the source to support the construction of the graph. We validate \method~in experiments with text-based games that require common sense and show that it outperforms baselines by 17\% of scores. The code is available at \url{https://github.com/ibm/diffg-rl}
\end{abstract}

\section{Introduction}
Taking into account background knowledge as the context has always been an important yet challenging part of solving tasks that involve natural language. One illustrative example of such challenges is text-based games. Text-based games are computer games where game states and action spaces are represented in pure texts. To play them, players have to not only understand in-game texts correctly but also make appropriate action decisions from given options according to the context. Computational agents required to solve such games naturally arise in the form of natural language processing (NLP) systems trained with reinforcement learning (RL) algorithms. However, the intrinsic properties of text-based games such as partial observability, long-term dependencies, sparse reward signals, and large action spaces make it extremely challenging for RL agents to learn. Specifically, the chance of agents discovering optimal actions from the vast combinatorial action spaces is astronomically low.

\begin{figure}[t]
  \begin{center}
  \includegraphics[width=\linewidth]{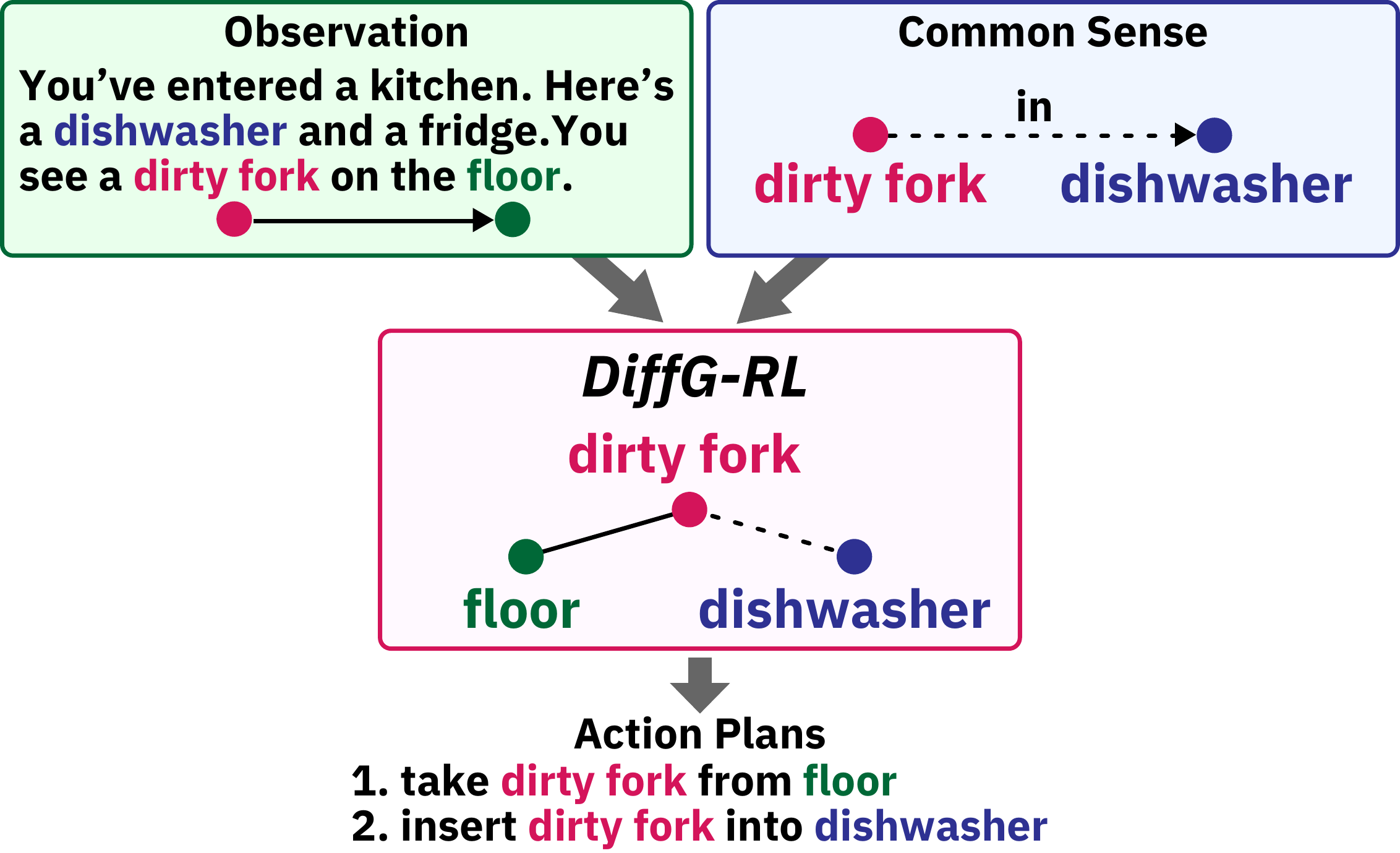}
  \end{center}
  \caption{An example of text-based games. Our proposed \method~summarizes the difference between the environment states and common sense in a graph and uses it as a basis for predicting a plan of action.}
  \vspace{-10pt}
  \label{fig:teaser}
\end{figure}

Previous approaches~\cite{transfer,efficient,wordplay,twc,kimura2021reinforcement,aaaiw} have used external knowledge to constraint agents' action outputs in order to shrink the size of search space. Recently, \citet{twc} and \citet{aaaiw} utilized human common sense which improved sample efficiency and enabled agents to perform look-ahead planning.
However, these approaches have not yet clarified how common sense should actually be used. Specifically, a huge amount of common sense is given at once, regardless of the environment states, and the correspondence between the states and common sense is unclear. This prevents agents from learning which common sense to use in which state, and the agents remember only the results after using common sense.

In this paper, as usage of common sense, we assume that differences between environment states and common sense can provide a basis for action selection and further improve sample efficiency.
For example, consider the environment state of ``dirty fork on the floor'' and the common sense of ``dirty fork should be in the dishwasher''. The difference between floor and dishwasher in the location of the dirty fork helps agents plan their actions to pick it up from the floor and put it in the dishwasher.
We construct a difference graph that maps environment states to common sense and explicitly represents their differences. Further, we develop an encoder dedicated to this graph and propose an agent that can effectively concentrate on learning which common sense to use in which state. An example is shown in Fig.~\ref{fig:teaser}.

Inevitable problems in constructing the difference graph are extracting the right amount and unifying the representation. For the first problem, large amounts of common sense cannot be encoded, and small amounts are insufficient for learning. In prior work, the amount of common sense is reduced by extracting only common sense that contains representations that exactly match the objects that appear in the environment, but in some tasks, there may be no common sense available due to mismatches of representations. In contrast, we extract the appropriate amounts of common sense based on semantics and the circumstances, independent of linguistic representations. For the second problem, to help agents recognize the difference between common sense and environmental states, their representations should be aligned. Therefore, we propose an extraction framework for common sense that includes an acquisition of appropriate amounts based on meanings and circumstances and a representation transformation that facilitates the mapping to environment states.

Our contributions in this work are as follows. (1) We introduce a difference graph with an explicit representation of the difference between the environment states and common sense and a novel agent with a dedicated graph encoder. (2) We develop a framework for extracting common sense from sources to facilitate comparing the environment states with common sense. (3) We perform experiments with text-based games and demonstrate that our approach outperforms baselines, and evaluate the effect of each component in our approach through ablation studies.

\section{Background}
\noindent\textbf{Text-based Games}: Text-based games can be formally framed as partially observable Markov decision processes (POMDPs), represented as a 7-tuple of $\langle  S, T, A, \Omega, O, R, \gamma \rangle $ denoting the set of environment states, conditional transition probabilities, actions, observations, conditional observation probabilities, reward function, and discount factor. We target choice-based games, where the player receives a textual observation $o_t \in \Omega$ and sends a short textual phrase from action choices $A$ to the environment as an action $a_t$. Most text-based games contain entities ($e_1, e_2, ..., e_M \in E_t$) such as items and location, and players often take actions on themselves (``go east'') or on items (``take dirty fork'').

\noindent\textbf{Common Sense and Text-based Games}: Given common sense, an agent receives an observation $o_t$ to determine the next action by comparing it with the common sense. Common sense is represented by an external knowledge graph stored as triplets of $\langle \textrm{subject}, \textrm{relationship}, \textrm{object} \rangle $, which is called a common sense graph. While the recently proposed TWC agent~\cite{twc} uses multiple graphs obtained from ConceptNet~\cite{conceptnet} and combines them, \cite{aaaiw} showed that just a single graph can suffice if Visual Genome (VG)~\cite{visualgenome} is used, as it contains more grounded graphs. We therefore use VG as a common sense source. There are two challenges when it comes to using VG with common sense: first, how to extract common sense from the source, and second, how agents use the common sense. In this work, we propose two methods to individually address these challenges.

\begin{figure*}[t]
  \begin{center}
  \includegraphics[width=\linewidth]{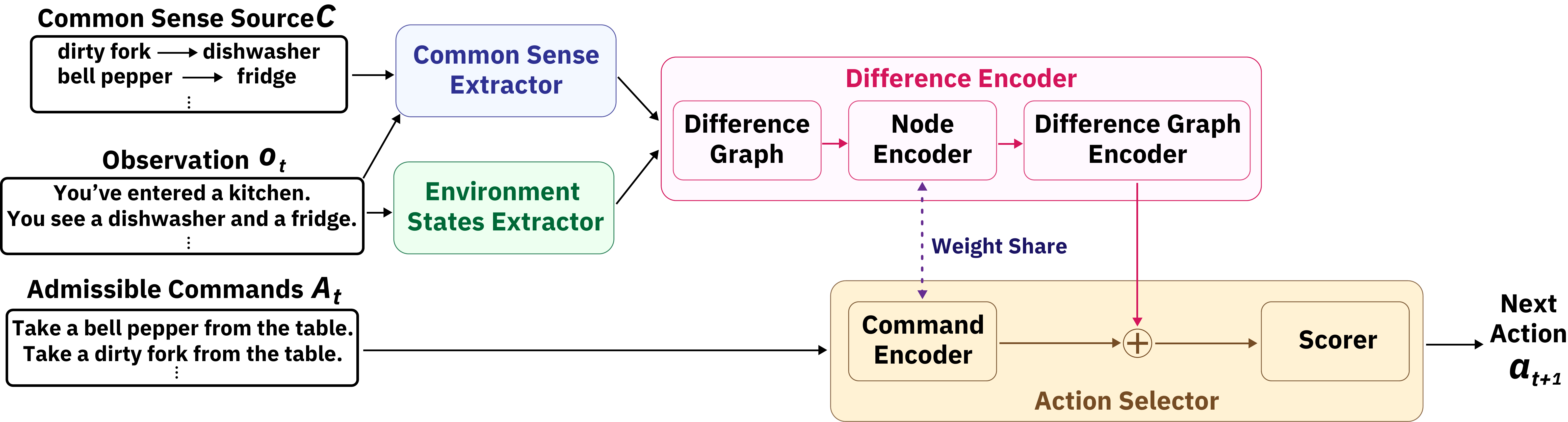}
  \end{center}
  \caption{An overview of our proposed \method~for text-based games based on the difference between the environment states and common sense. The Model consists of four components: Environment States Extractor, Common Sense Extractor, Difference Encoder, and Action Selector.}
  \label{fig:architecture}
\end{figure*}

\noindent\textbf{Environment States Extractor}: In this study, the current states of the environment are extracted from the observation $o_t$. Since states in text-based games have a graph-like structure, prior works~\cite{kgdqn,kga2c,efficient,gata} have represented environment states as a dynamic knowledge graph. KG-A2C introduced interactive objects ($io_1, io_2, ..., io_P \in I_t$), which are items that allow agents to interact directly with the surrounding environment. We connect these interactive objects to their state nodes (e.g., the locations they are in) in the knowledge graph and then separate them from the entities ($I_t \subset E_t$). We build on this to extract the states of interactive objects from observation text by using the Abstract Meaning Representation (AMR) parser~\cite{amr}. Nodes connected to the interactive objects by positional relationships (”on,” “in,” Etc.) from AMR are extracted as the current states. We also add a node representing the player "You" and use it when the interactive objects are in the inventory. At each step, the entities $E_t$, the interactive objects $I_t$, and their current states are updated on the basis of the observations, and the states of all interactive objects are tracked.
\label{sec:background}

\section{\method}
In this section, we first present an overview of our proposed agent, called \method, and then propose a framework for the extraction of common sense to facilitate the comparison of differences between environment states and common sense. Finally, we introduce the Difference Graph and its encoder, which provides a representation of the difference.

\subsection{Model Overview}
Our \method~agent with common sense for text-based games consists of the four components shown in Fig.~\ref{fig:architecture}.
\begin{enumerate}
  \item Environment States Extractor (EE): A component that extracts interactive objects and their current states from the observation texts $o_t$. The details were described in Section~\ref{sec:background}.
  \item Common Sense Extractor (CE): A framework for extracting the appropriate amount and representation of common sense to facilitate comparisons of the environment states and common sense.
  \item Difference Encoder (DE): A graph encoder and a node encoder of a difference graph that organize the environment states and common sense for interactive objects.
  \item Action Selector (AS): A component that includes an encoder of admissible commands $A_t$ and a selector that determines the next action $a_{t+1}$ from the output of the difference graph encoder and the command encoder.
\end{enumerate}

\subsection{Common Sense Extractor}
\label{sec:commonsenseextractor}
Giving the common sense all at once (e.g., all the triples of graphs in ConceptNet) is excessive and inefficient for solving the tasks. We have also not identified any way to encode it all at once so far. Existing research~\cite{twc, aaaiw} has narrowed it down to triples of common sense graphs ($c_1, c_2, ..., c_N \in C$) that perfectly match the entities ($e_1, e_2, ..., e_M \in E$) in text-based games, but common sense rarely has such a representation. 
We extract common sense based on word meaning and the circumstances of games, independent of linguistic representation, to broaden the scope of common sense extraction.
In addition, the difference graph described in Section~\ref{sec:difference} needs unified representations of environment states and common sense to help agents understand the correspondences between them. To extract the appropriate amount and representation of common sense, we propose a framework consisting of three components: extracting by meaning (EbM), narrowing by circumstances (NbC), and transforming into grounded representation (TGR).

\subsubsection{Extracting by Meaning}
\label{sec:extractingbymeaning}
As the first step to extract triples of common sense graphs by meaning, we utilize the similarity between vectors obtained by word embedding instead of spell matching. The similarity $sim$ is represented as

\begin{equation}
  \label{eq:similarity}
  sim(s_i, e_j) = \frac{\bm{s_i}\cdot\bm{e_j}}{|\bm{s_i}||\bm{e_j}|},
\end{equation}

where $s_i$ is a subject in a triple of common sense graph $c_i$, and $e_j$ represents an entity in text-based games. The bolded terms also represent vectors obtained by word embedding. If the similarity is greater than a preset threshold, it is considered to have a similar meaning. We calculate this for all combinations and then replace $s_i$ with an object in the triple of common sense graph and calculate them again. If both $s_i$ and $o_i$ are similar to one of $E$, its triple of common sense graph $c_i$ is extracted.

Note that because this component relaxes the constraints on common sense much more than with exact matching, the number of extracted triples of common sense graphs will be enormous.~\footnote{We tried doing the extraction with an NVIDIA TITAN X (Pascal) with 12 GB of memory, but the triples extracted from VG overwhelmed the available memory.} In most cases, it is necessary to use it in combination with the twc components introduced in \ref{sec:nbc} and \ref{sec:transforming}.

\subsubsection{Narrowing by Circumstances}
\label{sec:nbc}
We leave only triples of graphs that are in line with the circumstances of games, i.e. ``interactive object $\rightarrow$ object's state''. In many text-based games, ``interactive object $\rightarrow$ location'' (e.g., dirty fork $\rightarrow$ dishwasher) remains, while ``location $\rightarrow$ location'' (agents do not move the dishwasher into the fridge) etc. is removed.

\subsubsection{Transforming into Grounded Representation}
\label{sec:transforming}
We transform the subject and object in the extracted triple of common sense graph into the entities to which they correspond in the first component, the EbM  (Section~\ref{sec:extractingbymeaning}). In the case of Eq.~\ref{eq:similarity}, $s_i$ is transformed into $e_j$. This eliminates the influence of differences between the extracted common sense and games' representations and clarifies the correspondence between the environment states and common sense.

\subsection{Difference Encoder}
\label{sec:difference}
\subsubsection{Difference Graph}
\begin{figure}[t]
  \begin{center}
  \includegraphics[width=0.8\linewidth]{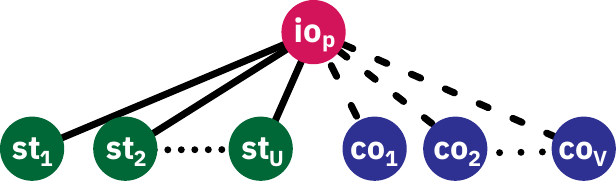}
  \end{center}
  \caption{Concept of the difference graph. \textcolor[cmyk]{0.212, 0.968, 0.479, 0}{Pink} nodes, \textcolor[cmyk]{0.890, 0.486, 0.998, 0.128}{green} nodes, and \textcolor[cmyk]{0.943, 0.920, 0.068, 0}{blue} nodes represent interactive objects, environment states, and common sense, respective.}
  \label{fig:graph}
\end{figure}

We introduce the difference graph to represent the difference between the environment states and common sense to select common sense according to the states and to obtain the basis for the next action plan, as shown in Fig.~\ref{fig:graph}. We define the difference graph as a representation of the situation where “an interactive object should be placed at A based on common sense, but is now placed at B” (a dirty fork should be placed at the dishwasher but is currently on the floor). The outputs of the current state extractor and the common sense extractor are organized by interactive objects ($io_1, io_2, ..., io_P \in I_t$). The difference graph contains three types of nodes, with multiple current state nodes ($st_1, st_2, ..., st_U \in \mathcal{U} (p)$) and common sense nodes ($co_1, co_2, ..., co_V \in \mathcal{V} (p)$) corresponding to one interactive object node ($io_p$). For edges, there are two types: interactive object-current state and interactive object-common sense. After the TGR  (Section~\ref{sec:transforming}), the common sense node has the same representation as the entities in games. The difference graph is updated in accordance with the observation texts at each time step.

\subsubsection{Node Encoder}
We convert the words in a node of the difference graph into a series of vectors by word embedding and obtain a fixed-length vector using the node encoder. We use the fixed-length vector as the initial feature of each node in the difference graph encoder. We use bidirectional GRU~\cite{gru} for the node encoder.

\subsubsection{Difference Graph Encoder}
\label{sec:dge}
We develop a graph encoder to encode the difference graph. Similar to recent graph neural networks, we update the features of a node by aggregating the features of its neighbors. The aggregate of our encoder is based on the Graph Isomorphism Network (GIN)~\cite{gin} and is calculated as 
\begin{multline}
  h_{io_p}^{(k)} = \textrm{MLP}\{ \phi (1 + W_I) h_{io_p}^{(k-1)} \\
  + \sum_{st_u \in \hspace{2pt} \mathcal{U} (p)} \phi (W_{ST}h_{st_u}^{(k-1)}) \\
  + \sum_{co_v \in \mathcal{V} (p)} \phi (W_{CO}h_{co_v}^{(k-1)}) \},
  \label{eq:dge}
\end{multline}
where $h_X^{(k)}$ represents the feature of $X$ node with $k$ iterations of the aggregation, $\phi$ represents an activation function, and $\textrm{MLP}$ represents multi-layer perceptrons.

To distinguish between the three types of node and represent the difference between the current state and common sense, different learnable parameters are provided for each type: $W_I$, $W_{ST}$, and $W_{CO}$. Since the actions are based on the interactive objects, the encoder only aggregates for the interactive object $io_p$. In GIN, one MLP is used after the product with the learnable parameter because MLPs can represent a composition of functions, but we add an activation function $\phi$ for output simplification and training stability.

The aggregation can be repeated to reflect the features of distant nodes. As a results, the difference graph encoder can handle the environment states and common sense even if they become subgraphs consisting of multiple nodes. Note that we assume the maximum distance of 1 from the interactive object in the following experiments.

\subsection{Action Selector}
The action selector calculates the probability of each action from the concatenation of the vector representation $\bm{a_t^i}$ of the admissible command $a_t^i \in A_t$ and the output $\bm{d_t}$ of the difference graph encoder. $\bm{a_t^i}$ is obtained by word embedding and the command encoder, similar to the node encoder in the difference graph encoder. We use the bidirectional GRU for the command encoder and share weight with the node encoder. The scorer consists of two MLP layers, a dropout layer, and an activation layer and calculates the probability $p_{a_t^i} = \textrm{Scorer}(\bm{a_t^i}, \bm{d_t})$.

\section{Experiments}

\subsection{Environment}
\begin{table}[t]
  \small
  \centering
  \caption{Specifications of TWC game that we use.}
  \begin{tabular}[h]{>{\centering}m{0.2\linewidth}>{\centering}m{0.4\linewidth}>{\centering\arraybackslash}m{0.15\linewidth}}
      \toprule
      Level & Interactive objects & Rooms \\
      \midrule
      Easy & 1 & 1 \\
      Medium & 3 & 1 \\
      Hard & 7 & 2 \\
      \bottomrule
  \end{tabular}
  \label{tab:twc}
\end{table}
We conduct experiments with the TWC game~\cite{twc} to verify the difference between common sense and the environment states. The goal of the TWC game is to tidy up a house by putting items where they should be and requires common sense about the relationships between objects and their locations. We generate a new game set using the scripts from the original TWC study~\cite{twc}. There are three difficulty levels depending on the number of rooms and the number of interactive objects, as shown in Tab.~\ref{tab:twc}. Because the agent performance is affected by the number of objects/rooms, we unify the different numbers of these included in the same difficulty level in the original dataset. 

To test the generalization performance, we introduce a supervised learning paradigm and split the dataset into three subsets: train, test, and valid. The original dataset~\cite{twc} contains only five games each in the train and test sets, but we generate 100 games and split them into $\textrm{train} : \textrm{test} : \textrm{valid} = 50 : 40 : 10$. 

The TWC game contains two test sets: an \textit{IN} set with the same entities as the train set and an \textit{OUT} set consisting of entities that do not appear in the train set. The \textit{OUT} set cannot be solved simply by memorizing the results (i.e., pairs of interactive objects and locations) after using common sense. 
Depending on the situation, such as the type of room the agents are in now, the same thing may be placed in different places between the train set and the \textit{OUT} set.
Therefore, the \textit{OUT} set is used for the validation and ablation study because it is suitable for evaluating the ability to use common sense in a given situation.

\begin{table*}[t]
  \centering
  \small
  \caption{General results for two test games: \textit{IN} within the training distribution of entities and \textit{OUT} outside the distribution. All experiments were performed with five random seeds. Each value is a pair $(\textrm{average}) \pm (\textrm{standard deviation})$. We highlight the \textbf{best} model in bold. }
  \begin{tabular}[h]{>{\centering}m{0.005\linewidth}>{\raggedright}m{0.4\linewidth}>{\centering}m{0.1\linewidth}>{\centering}m{0.1\linewidth}>{\centering\arraybackslash}m{0.1\linewidth}}
    \toprule
      & Method & Easy & Medium & Hard \\
     \midrule
     \multirow{5}{*}{\rotatebox{90}{\textit{IN}}} & KG-A2C~\cite{kga2c} & 0.89 $\pm$ 0.02 & 0.76 $\pm$ 0.02 & 0.33 $\pm$ 0.01 \\
      & TWC agent-CN~\cite{twc} & 0.91 $\pm$ 0.02 & 0.75 $\pm$ 0.02 & 0.31 $\pm$ 0.01 \\
      & TWC agent-VG~\cite{aaaiw} & 0.92 $\pm$ 0.01 & 0.69 $\pm$ 0.03 & 0.32 $\pm$ 0.02 \\
      & TWC agent-VG+KG-A2C & \textbf{0.95 $\pm$ 0.01} & \textbf{0.82 $\pm$ 0.02} & 0.26 $\pm$ 0.01 \\
      & \textbf{\method} & \textbf{0.95 $\pm$ 0.00} & \textbf{0.82 $\pm$ 0.02} & \textbf{0.38 $\pm$ 0.02} \\
     \midrule
     \multirow{5}{*}{\rotatebox{90}{\textit{OUT}}} & KG-A2C~\cite{kga2c} & 0.78 $\pm$ 0.03 & 0.72 $\pm$ 0.02 & 0.33 $\pm$ 0.01 \\
      & TWC agent-CN~\cite{twc} & 0.77 $\pm$ 0.03 & 0.69 $\pm$ 0.02 & 0.29 $\pm$ 0.02 \\
      & TWC agent-VG~\cite{aaaiw} & 0.78 $\pm$ 0.03 & 0.67 $\pm$ 0.02 & 0.25 $\pm$ 0.02 \\
      & TWC agent-VG+KG-A2C & 0.82 $\pm$ 0.03 & 0.72 $\pm$ 0.02 & 0.25 $\pm$ 0.01 \\
      & \textbf{\method} & \textbf{0.91 $\pm$ 0.04} & \textbf{0.76 $\pm$ 0.02} & \textbf{0.35 $\pm$ 0.02} \\
     \bottomrule
  \end{tabular}
  \label{tab:generalresults}
\end{table*}

\subsection{Methods and Metrics}
We use four baselines.
\begin{itemize}
  \item KG-A2C~\cite{kga2c} is a method with our implementation that organizes the environment states obtained from observation texts into a knowledge graph.
  \item TWC~agent-CN~\cite{twc} is a method that uses common sense obtained from ConceptNet.
  \item TWC~agent-VG~\cite{aaaiw} is a method that utilizes the same model as TWC~agent-CN but with VG as the common sense source.
  \item TWC~agent-VG+KG-A2C is a method that naively combines the TWC~agent-VG and KG-A2C without our proposed difference encoder and common sense extractor.
\end{itemize}

We evaluate performance on the normalized score computed by dividing the actual score by the maximum possible score. The scores range from 0 to 1, and higher is better.

\subsection{Implementation and Training Details}
In all methods, we use GloVe~\cite{glove} provided by GENSIM~\footnote{https://radimrehurek.com/gensim/} for word embedding. The hidden size is set to 512 for \method~and to 300 for the baselines. The activation function is ELU~\cite{elu} for \method~and ReLU~\cite{relu} for the baselines.
The threshold of similarity for the EbM (Section~\ref{sec:extractingbymeaning}) is set to 0.3.

For training, we optimize all models for 100 epochs with Adam~\cite{adam} optimizer using a learning rate of $3.0\times10^{-5}$ and the default hyperparameters in PyTorch~\cite{pytorch}.
For tests, the model with the largest normalized score and the smallest number of steps in the validation is used. \method~can train 100 epochs in two and a half days using a single NVIDIA TITAN X (Pascal) GPU.

\subsection{General Results}
\begin{figure}[t]
  \begin{center}
  \includegraphics[width=0.95\linewidth]{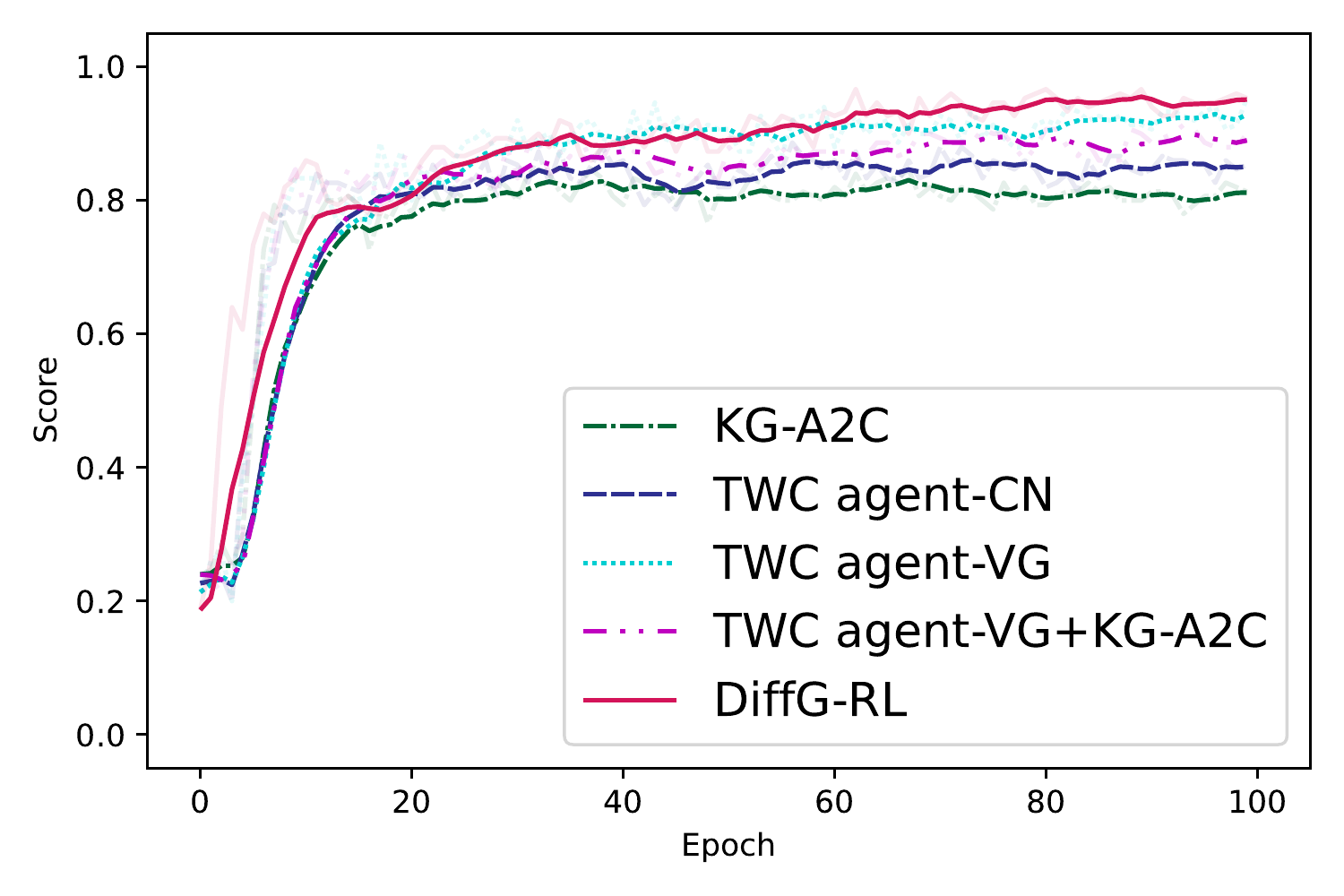}
  \end{center}
  \caption{Performance evaluation for the medium level in the training. (Smoothing is performed to clarify the differences in the results of a single run.)}
  \label{fig:training}
\end{figure}

\begin{table*}[t]
  \centering
  \small
  \caption{Comparison of performance to evaluate the components, EbM, NbC, TGR (Section~\ref{sec:commonsenseextractor}), and DE (Section~\ref{sec:difference}) in \method. We use hard level games in \textit{OUT}.}
  \begin{tabular}[t]{>{\centering}m{0.03\linewidth}>{\centering}m{0.04\linewidth}>{\centering}m{0.04\linewidth}>{\centering}m{0.04\linewidth}>{\centering}m{0.04\linewidth}>{\centering}m{0.2\linewidth}>{\centering}m{0.15\linewidth}>{\centering\arraybackslash}m{0.1\linewidth}}
    \toprule
    No. & EbM & NbC & TGR & DE & Precision & Recall & Scores \\
    \midrule
    1 & & & & & 23 / 896 (2.6\%) & 23 / 257 (6.4\%) & 0.25 $\pm$ 0.02 \\
    2 & \checkmark & & & & 2.80M / 11.00M (25.6\%) & 343 / 357 (96.1\%) & N/A \\
    3 & \checkmark & \checkmark & & & 2.80M / 6.05M (37.1\%) & 343 / 357 (96.1\%) & N/A \\
    4 & \checkmark & \checkmark & \checkmark & & 343 / 5414 (6.3\%) & 343 / 357 (96.1\%) & 0.26 $\pm$ 0.01 \\
    5 & \checkmark & \checkmark & \checkmark & \checkmark & 343 / 5414 (6.3\%) & 343 / 357 (96.1\%) & \textbf{0.35 $\pm$ 0.02} \\
    \bottomrule
  \end{tabular}
  \label{tab:componentsresults}
\end{table*}

\begin{table}[t]
  \small
  \centering
  \caption{Ablation results for the relationship between the similarity threshold (TH) in EbM and the extracted common sense graphs. CE denotes the common sense extractor (Section~\ref{sec:commonsenseextractor}). We use goal graphs of hard level games in the \textit{OUT} set.}
  \begin{tabular}[t]{>{\centering}m{0.02\linewidth}>{\centering}m{0.06\linewidth}>{\centering}m{0.03\linewidth}>{\centering}m{0.32\linewidth}>{\centering\arraybackslash}m{0.3\linewidth}}
    \toprule
    No. & CE & TH & Precision & Recall \\
    \midrule
    1 & & - & 23 / 896 (2.6\%) & 23 / 357 (6.4\%) \\
    2 & \checkmark & 0.6 & 268 / 3707 (7.2\%) & 268 / 357 (75.1\%) \\
    3 & \checkmark & 0.5 & 321 / 2984 (10.8\%) & 321 / 357 (89.9\%) \\
    4 & \checkmark & 0.4 & 341 / 5151 (6.6\%) & 341 / 357 (95.5\%) \\
    5 & \checkmark & 0.3 & 343 / 5414 (6.3\%) & 343 / 357 (96.1\%) \\
    \bottomrule
  \end{tabular}
  \label{tab:componentsextraction}
\end{table}

Table~\ref{tab:generalresults} lists the results of the \textit{IN} and \textit{OUT} test sets achieved by the baselines and the proposed approach (\method) trained for each difficulty level. \method~using the difference graph outperforms the baselines in all results. Specifically, \method~improves 40\% in the hard level using the \textit{OUT} set from TWC~agent-VG, which uses only common sense, and 17\% in the easy level using the \textit{OUT} set from KG-A2C (previous SOTA), which uses only a knowledge graph.
We can observe high performances on \textit{OUT}, which cannot be solved by simply memorizing the results after using common sense in the training. This suggests that the representation of the difference between the environment states and common sense in the proposed approach contributes to learning how to use common sense.

Table~\ref{tab:generalresults} also shows that TWC~agent-VG+KG-A2C struggles on the hard difficulty level and is less than or equal to KG-A2C and TWC~agent-VG, which use only a knowledge graph of the environment states or common sense (not both). We believe this approach is vulnerable to an increase in the environment states and common sense as the number of interactive object increases. In contrast, \method~shows a solid improvement over KG-A2C and TWC~agent-VG, indicating that it is able to effectively utilize a combination of the environment states and common sense to deal with increased interactive objects.

Figure~\ref{fig:training} shows the training curves at the medium level for the baselines and \method, where it is clear that \method~performs the best. We believe the difference between the environment states and common sense has a positive impact on the decision making and improves the sample efficiency.

Considering both Tab.~\ref{tab:generalresults} and Fig.~\ref{fig:training} together, in the medium level games, we can see that TWC~agent-VG has a low performance on both test sets, despite its high performance in the training. In contrast, \method~performs well on all of the training and the two test sets. 
This reinforces our intuition that the difference graph of our proposed approach improves the generalization performance.

\subsection{Ablation Study}

\begin{figure*}[t]
  \begin{center}
  \includegraphics[width=\linewidth]{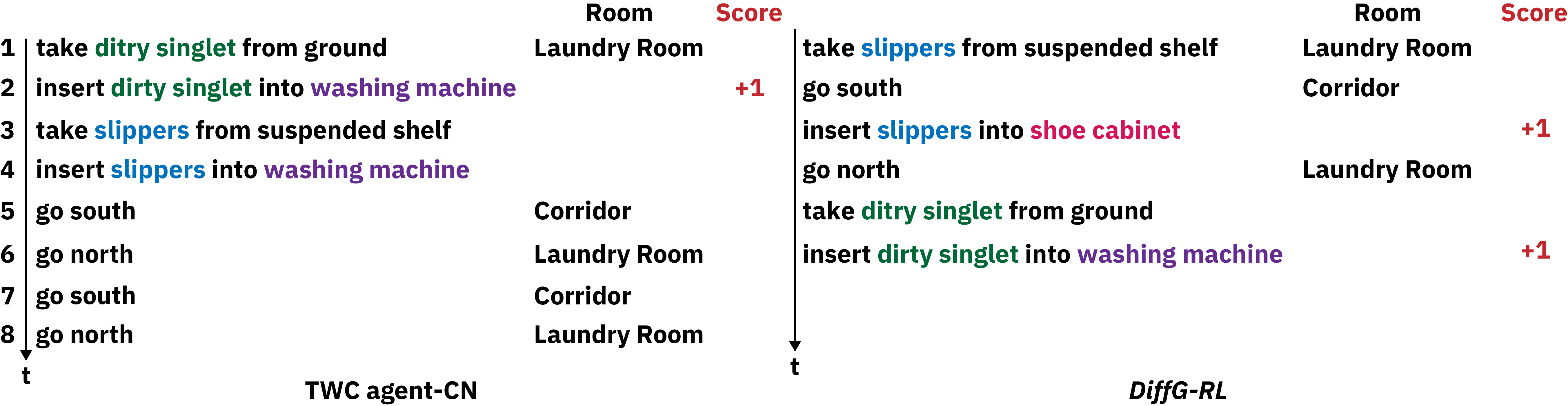}
  \end{center}
  \caption{An comparison of agents' behavior. TWC agent-CN (left) and \method~(right) performed the task of placing a \textcolor[cmyk]{0.890, 0.486, 0.998, 0.128}{dirty singlet} and \textcolor[cmyk]{0.859, 0.525, 0.055, 0}{slippers}, which is in the hard difficulty level and uses the \textit{OUT} set.}
  \label{fig:qualitative}
\end{figure*}

\noindent \textbf{Effect of Components}: 
We investigate the effect of the components in \method, and show the results in Tab.~\ref{tab:componentsresults}. Based on the field of information retrieval, we add precision and recall to the metric to evaluate how well the extraction method can extract the common sense needed to achieve the game goals. We first introduce the concept of a goal graph $g_1, g_2, ..., gL \in G$ directly connecting interactive objects and their goal locations. The precision is calculated by dividing the number of triples of common sense graphs that correspond to the triple of goal graph $C_g \subset C$ by the total number $N$ of $C$. The recall is calculated by dividing the number of triples of goal graphs that are covered by the triples of common sense graphs $G_c \subset G$ by the total number $L$ of $G$.

We believe that an agent's performance can be improved by giving it a computationally feasible number of commonsense knowledge triples, which cover the common sense needed to solve the problem. Thus, higher both precision and recall are better, but there is a tradeoff between the two. We also use TWC~agent-VG+KG-A2C as the most naive method (no.1), and since the performance difference between the proposed method and no.1 is the largest in Tab.~\ref{tab:generalresults}, we use the hard level games in the \textit{OUT} set for the score.

We can see that using EbM significantly improves the recall. This reinforces the effectiveness of our idea that common sense should be extracted by meaning independent of linguistic representation. We also see that the precision is also greatly improved by using NbC together. However, the number of triples of common sense graphs extracted is still huge, as indicated by the denominator values of the precision in no.3. No.2 and no.3 could not be executed because the number of triples exceeds the available GPU memory. Since TGR unified the representation of goal graphs and the extracted common sense graphs, multiple triples can be combined into one and the number of the extracted triples can be reduced. Therefore, we believe that the proposed components should not be used individually but as a framework that brings them together.

We also see from the difference in scores that both CE and DE contribute to the performance improvement. Comparing their respective contributions, we can see that DE has a greater impact on performance than the other components (4 $\rightarrow$ 5). This leads us to believe that the difference graph is relatively critical in \method~for its impact on the agents' decision-making.

\noindent \textbf{Similarity Threshold}: 
We investigate the relationship between the similarity threshold in EbM and the number of the extracted triples of common sense graphs. We compute the precision and the recall between the extracted common sense graphs and the goal graphs in the same way as in Tab.~\ref{tab:componentsresults} and show the results in Tab.~\ref{tab:componentsextraction}. We can observe that the precision and the recall are a tradeoff tendency when the threshold exceeds 0.5. To maximize the performance of DE by providing common sense graphs that correspond to the goal graphs in all games, we focus on the recall in our experiments and set the threshold at 0.3. Note that lowering the threshold increases the noise (the precision decreases), so a method to increase the recall while keeping the threshold high will be required in the future (see Section~\ref{sec:limitations}).

\subsection{Qualitative Results}
Figure~\ref{fig:qualitative} shows a comparison of the agent's behavior between TWC~agent-CN and \method. The tasks are to put the dirty singlet in the laundry room (north) into the washing machine and to put the slippers in the suspended shelf in the laundry room into the shoe cabinet in the corridor (south).

TWC~agent-CN puts the dirty singlet into the washing machine correctly, but it also puts the slippers into the washing machine wrongly. We assume that this is because by giving the agent a single vector that aggregates common sense knowledge triples about all interactive objects, the agent has been strongly affected by common sense for the dirty singlet. We also observe that for slippers where the goal exists in another room, it repeats the move commands (such as ``go south'' and ``go north'') and does not reach the correct location. This can be considered an incorrect understanding of the current state. However, \method~can put the three interactive objects back where they should be in order, even if it requires moving the room. We believe \method~is robust to such effects of common sense about other interactive objects because it explicitly encodes the correspondences between the current location and common sense for each interactive object using a difference graph. 

\section{Related Work}
\noindent \textbf{Common Sense for Text-based Games}: 
Many recent methods have focused on providing common sense to agents to efficiently explore the vast observation and action spaces of text-based games. 
\citet{twc} proposed a text-based game TextWorld Commonsense (described as the TWC game in this paper) that requires common sense from agents and a baseline TWC agent that utilize common sense obtained from ConceptNet~\cite{conceptnet}. \citet{aaaiw} proposed using scene graph datasets such as VG~\cite{visualgenome} as a more grounded common sense source base on the TWC agent.
BiKE~\cite{efficient} shares information between the state graph and the common sense graph by means of a bidirectional attention mechanism, but focuses only on information that is similar between nodes.
\citet{transfer} transfer common sense trained in other games into the target game strategy. \citet{wordplay} utilize common sense obtained in large-scale models such as COMET~\cite{comet} and BERT~\cite{bert} based on KG-A2C~\cite{kga2c}. Common sense also has a high affinity with logic rules, and some studies~\cite{kimura2021reinforcement, kimura-etal-2021-loa, kimura-etal-2021-neuro, chaudhury-etal-2021-neuro} combine them to solve text-based games.
Since these methods do not clarify the correspondence between the environment states and common sense, it is difficult to learn how to use common sense according to the situation.

In contrast, our work extracts common sense by meaning and graphically represents the difference between common sense and the environment states.

\noindent \textbf{Knowledge Graph Extraction}: 
Some prior works have utilized state representations using knowledge graphs to effectively prune the vast observation and action space of text-based games. KG-DQN~\cite{kgdqn} and KG-A2C use OpenIE~\cite{openie} to create a knowledge graph of the game's belief state from observation texts. GATA~\cite{gata} uses self-supervised learning to train the construction and update the belief graph. Worldfromer~\cite{worldformer} uses the world models to simultaneously tasks agents with a next-step belief graph prediction as well as the usual action generation. These methods utilize knowledge graph representation of the environment states, but do not use prior knowledge such as common sense.

Our work differs in that we utilize a knowledge graph to represent not only the environment states but also the correspondence with common sense in the form of the different graphs.

\noindent \textbf{Usage of Common Sense}
Extraction of suitable commonsense statements is well-studied~\cite{ma-etal-2019-towards, lin-etal-2019-kagnet}, and recall of commonsense knowledge for a task receives much attention~\cite{lin-etal-2019-kagnet, ilievski2020consolidating} in QA. However, as mentioned in 1, we focus on the denominator of precision as well as recall, which we believe is a new perspective. In addition, existing extraction methods~\cite{ma-etal-2019-towards, lin-etal-2019-kagnet} use triples with matched words. However, in this study, we use triples that are close in the distance between vectors after transformation by word embedding. 

The filtering of common sense statements based on common sense is also studied in works related to the affordance of objects or defeasible reasoning~\cite{qasemi2022paco, rudinger-etal-2020-thinking, do-pavlick-2021-rotten}. They argue that the performance of state-of-the-art language models drops significantly in updating inferences when the context changes from the general situations by using their original tasks and datasets. In contrast, this paper proposes a model that explicitly represents the dynamically changing context of environmental states and common sense differences in RL and argues for its effectiveness through experiments. 

\section{Conclusion}
In this work, we investigated the difference between the environment state and common sense as a basis for RL agent decision-making in text-based games. We proposed \method, a novel agent that constructs a graph that represents the difference, along with a dedicated encoder, also contains a common sense extraction framework to obtain the appropriate amount and representation of common sense to facilitate the comparison between the environment states and the common sense. Our experimental results showed that \method~outperformed baselines that used only a knowledge graph, only common sense, or a naive combination of the two. These findings demonstrate the effectiveness of the difference graph, which is a representation of the difference between the environment states and common sense, for text-based games.

\section{Limitations}
\label{sec:limitations}
An important aspect of our approach is that it utilizes the difference between the environment states and common sense as the basis for decision-making. However, the TWC game does not consider the relationships with other objects, which means the agents were sometimes not provided sufficient context. For example, it is difficult for an agent to determine whether a dirty fork should be placed in a dishwasher or on a dining table based solely on the information that it is holding a dirty fork in its hand. The location depends on further contexts, such as whether there is food left on the plates or whether the person eating is full. Since agents can only repeat their attempts based on scores in the TWC game, we hope to validate agents in games that provide more context, such as ALFWorld~\cite{ALFWorld20} and ScienceWorld~\cite{scienceworld2022}. Our approach should be able to support such high-context situations if we extend the representation of the environment states in the difference graph from node to sub-graph (details are provided in Section~\ref{sec:dge}).

In addition, although we used GloVe for word embedding, this is slightly outdated considering the recent development of natural language processing. The performance of our approach could be further improved by using more up-to-date word embedding. This may also allow us to obtain a sufficient number of graphs even if we raise the similarity threshold, which was set to 0.3 in our experiments.

\section{Broader Impact}
Our model does not use sensitive contexts such as legal or medical data. In addition, the dataset and common sense sources used in our experiments do not contain sensitive information. Since the actions taken by agents in the proposed model are based on the difference between the environment states and common sense, we can analyze the difference to reveal the reasons behind the actions. For example, if the model is biased, it can help to identify the cause of the biased behavior. However, when adding new common sense, it is necessary to thoroughly examine the model for bias, including that which has already been added.

\bibliography{anthology, custom}

\begin{thebibliography}{34}
\expandafter\ifx\csname natexlab\endcsname\relax\def\natexlab#1{#1}\fi

\bibitem[{Adhikari et~al.(2020)Adhikari, Yuan, C^^c3^^b4t^^c3^^a9, Zelinka,
  Rondeau, Laroche, Poupart, Tang, Trischler, and ~}]{gata}
Ashutosh Adhikari, Xingdi~(Eric) Yuan, Marc-Alexandre C^^c3^^b4t^^c3^^a9,
  Mikulas Zelinka, Marc-Antoine Rondeau, Romain Laroche, Pascal Poupart, Jian
  Tang, Adam Trischler, and William L.~Hamilton ~. 2020.
\newblock Learning dynamic belief graphs to generalize on text-based games.
\newblock In \emph{NeurIPS 2020}. ACM.

\bibitem[{Ammanabrolu and Hausknecht(2020)}]{kga2c}
Prithviraj Ammanabrolu and Matthew Hausknecht. 2020.
\newblock \href {https://openreview.net/forum?id=B1x6w0EtwH} {Graph constrained
  reinforcement learning for natural language action spaces}.
\newblock In \emph{International Conference on Learning Representations}.

\bibitem[{Ammanabrolu and Riedl(2019{\natexlab{a}})}]{kgdqn}
Prithviraj Ammanabrolu and Mark Riedl. 2019{\natexlab{a}}.
\newblock \href {https://www.aclweb.org/anthology/N19-1358} {Playing
  text-adventure games with graph-based deep reinforcement learning}.
\newblock In \emph{Proceedings of the 2019 Conference of the North {A}merican
  Chapter of the Association for Computational Linguistics: Human Language
  Technologies, Volume 1 (Long and Short Papers)}, pages 3557--3565,
  Minneapolis, Minnesota. Association for Computational Linguistics.

\bibitem[{Ammanabrolu and Riedl(2019{\natexlab{b}})}]{transfer}
Prithviraj Ammanabrolu and Mark Riedl. 2019{\natexlab{b}}.
\newblock \href {https://doi.org/10.18653/v1/D19-5301} {Transfer in deep
  reinforcement learning using knowledge graphs}.
\newblock In \emph{Proceedings of the Thirteenth Workshop on Graph-Based
  Methods for Natural Language Processing (TextGraphs-13)}, pages 1--10, Hong
  Kong. Association for Computational Linguistics.

\bibitem[{Ammanabrolu and Riedl(2021)}]{worldformer}
Prithviraj Ammanabrolu and Mark Riedl. 2021.
\newblock Learning knowledge graph-based world models of textual environments.
\newblock In \emph{Advances in Neural Information Processing Systems}.

\bibitem[{Angeli et~al.(2015)Angeli, Johnson~Premkumar, and Manning}]{openie}
Gabor Angeli, Melvin~Jose Johnson~Premkumar, and Christopher~D. Manning. 2015.
\newblock \href {https://doi.org/10.3115/v1/P15-1034} {Leveraging linguistic
  structure for open domain information extraction}.
\newblock In \emph{Proceedings of the 53rd Annual Meeting of the Association
  for Computational Linguistics and the 7th International Joint Conference on
  Natural Language Processing (Volume 1: Long Papers)}, pages 344--354,
  Beijing, China. Association for Computational Linguistics.

\bibitem[{Bosselut et~al.(2019)Bosselut, Rashkin, Sap, Malaviya,
  ^^c3^^87elikyilmaz, and Choi}]{comet}
Antoine Bosselut, Hannah Rashkin, Maarten Sap, Chaitanya Malaviya, Asli
  ^^c3^^87elikyilmaz, and Yejin Choi. 2019.
\newblock Comet: Commonsense transformers for automatic knowledge graph
  construction.
\newblock In \emph{Proceedings of the 57th Annual Meeting of the Association
  for Computational Linguistics (ACL)}.

\bibitem[{Chaudhury et~al.(2021)Chaudhury, Sen, Ono, Kimura, Tatsubori, and
  Munawar}]{chaudhury-etal-2021-neuro}
Subhajit Chaudhury, Prithviraj Sen, Masaki Ono, Daiki Kimura, Michiaki
  Tatsubori, and Asim Munawar. 2021.
\newblock \href {https://aclanthology.org/2021.emnlp-main.245} {Neuro-symbolic
  approaches for text-based policy learning}.
\newblock In \emph{Proceedings of the 2021 Conference on Empirical Methods in
  Natural Language Processing}, pages 3073--3078, Online and Punta Cana,
  Dominican Republic. Association for Computational Linguistics.

\bibitem[{Cho et~al.(2014)Cho, van Merri{\"e}nboer, Gulcehre, Bahdanau,
  Bougares, Schwenk, and Bengio}]{gru}
Kyunghyun Cho, Bart van Merri{\"e}nboer, Caglar Gulcehre, Dzmitry Bahdanau,
  Fethi Bougares, Holger Schwenk, and Yoshua Bengio. 2014.
\newblock \href {https://doi.org/10.3115/v1/D14-1179} {Learning phrase
  representations using {RNN} encoder{--}decoder for statistical machine
  translation}.
\newblock In \emph{Proceedings of the 2014 Conference on Empirical Methods in
  Natural Language Processing ({EMNLP})}, pages 1724--1734, Doha, Qatar.
  Association for Computational Linguistics.

\bibitem[{Clevert et~al.(2016)Clevert, Unterthiner, and Hochreiter}]{elu}
Djork-Arn^^c3^^a9 Clevert, Thomas Unterthiner, and Sepp Hochreiter. 2016.
\newblock Fast and accurate deep network learning by exponential linear units
  (elus).
\newblock In \emph{4th International Conference on Learning Representations,
  {ICLR} 2016}.

\bibitem[{Devlin et~al.(2019)Devlin, Chang, Lee, and Toutanova}]{bert}
Jacob Devlin, Ming-Wei Chang, Kenton Lee, and Kristina Toutanova. 2019.
\newblock \href {https://doi.org/10.18653/v1/N19-1423} {{BERT}: Pre-training of
  deep bidirectional transformers for language understanding}.
\newblock In \emph{Proceedings of the 2019 Conference of the North {A}merican
  Chapter of the Association for Computational Linguistics: Human Language
  Technologies, Volume 1 (Long and Short Papers)}, pages 4171--4186,
  Minneapolis, Minnesota. Association for Computational Linguistics.

\bibitem[{Do and Pavlick(2021)}]{do-pavlick-2021-rotten}
Nam Do and Ellie Pavlick. 2021.
\newblock \href {https://doi.org/10.18653/v1/2021.findings-acl.181} {Are rotten
  apples edible? challenging commonsense inference ability with exceptions}.
\newblock In \emph{Findings of the Association for Computational Linguistics:
  ACL-IJCNLP 2021}, pages 2061--2073, Online. Association for Computational
  Linguistics.

\bibitem[{Ilievski et~al.(2020)Ilievski, Szekely, Cheng, Zhang, and
  Qasemi}]{ilievski2020consolidating}
Filip Ilievski, Pedro Szekely, Jingwei Cheng, Fu~Zhang, and Ehsan Qasemi. 2020.
\newblock Consolidating commonsense knowledge.
\newblock \emph{arXiv preprint arXiv:2006.06114}.

\bibitem[{Keyulu et~al.(2019)Keyulu, Weihua, Jure, and Stefanie}]{gin}
Xu~Keyulu, Hu~Weihua, Leskovec Jure, and Jegelka Stefanie. 2019.
\newblock \href {https://openreview.net/forum?id=ryGs6iA5Km} {How powerful are
  graph neural networks?}
\newblock In \emph{International Conference on Learning Representations}.

\bibitem[{Kimura et~al.(2021{\natexlab{a}})Kimura, Chaudhury, Ono, Tatsubori,
  Agravante, Munawar, Wachi, Kohita, and Gray}]{kimura-etal-2021-loa}
Daiki Kimura, Subhajit Chaudhury, Masaki Ono, Michiaki Tatsubori, Don~Joven
  Agravante, Asim Munawar, Akifumi Wachi, Ryosuke Kohita, and Alexander Gray.
  2021{\natexlab{a}}.
\newblock \href {https://doi.org/10.18653/v1/2021.acl-demo.27} {{LOA}: Logical
  optimal actions for text-based interaction games}.
\newblock In \emph{Proceedings of the 59th Annual Meeting of the Association
  for Computational Linguistics and the 11th International Joint Conference on
  Natural Language Processing: System Demonstrations}, pages 227--231, Online.
  Association for Computational Linguistics.

\bibitem[{Kimura et~al.(2020)Kimura, Chaudhury, Wachi, Kohita, Munawar,
  Tatsubori, and Gray}]{kimura2021reinforcement}
Daiki Kimura, Subhajit Chaudhury, Akifumi Wachi, Ryosuke Kohita, Asim Munawar,
  Michiaki Tatsubori, and Alexander Gray. 2020.
\newblock Reinforcement learning with external knowledge by using logical
  neural networks.
\newblock In \emph{IJCAI-PRICAI-W}.

\bibitem[{Kimura et~al.(2021{\natexlab{b}})Kimura, Ono, Chaudhury, Kohita,
  Wachi, Agravante, Tatsubori, Munawar, and Gray}]{kimura-etal-2021-neuro}
Daiki Kimura, Masaki Ono, Subhajit Chaudhury, Ryosuke Kohita, Akifumi Wachi,
  Don~Joven Agravante, Michiaki Tatsubori, Asim Munawar, and Alexander Gray.
  2021{\natexlab{b}}.
\newblock \href {https://doi.org/10.18653/v1/2021.emnlp-main.283}
  {Neuro-symbolic reinforcement learning with first-order logic}.
\newblock In \emph{Empirical Methods in Natural Language Processing}, pages
  3505--3511. Association for Computational Linguistics.

\bibitem[{Kingma and Ba(2015)}]{adam}
Diederik~P. Kingma and Jimmy Ba. 2015.
\newblock \href {http://arxiv.org/abs/1412.6980} {Adam: {A} method for
  stochastic optimization}.
\newblock In \emph{3rd International Conference on Learning Representations,
  {ICLR} 2015, San Diego, CA, USA, May 7-9, 2015, Conference Track
  Proceedings}.

\bibitem[{Krishna et~al.(2017)Krishna, Zhu, Groth, Johnson, Hata, Kravitz,
  Chen, Kalantidis, Li, Shamma et~al.}]{visualgenome}
Ranjay Krishna, Yuke Zhu, Oliver Groth, Justin Johnson, Kenji Hata, Joshua
  Kravitz, Stephanie Chen, Yannis Kalantidis, Li-Jia Li, David~A Shamma, et~al.
  2017.
\newblock Visual genome: Connecting language and vision using crowdsourced
  dense image annotations.
\newblock \emph{International journal of computer vision}, 123(1):32--73.

\bibitem[{Lin et~al.(2019)Lin, Chen, Chen, and Ren}]{lin-etal-2019-kagnet}
Bill~Yuchen Lin, Xinyue Chen, Jamin Chen, and Xiang Ren. 2019.
\newblock \href {https://doi.org/10.18653/v1/D19-1282} {{K}ag{N}et:
  Knowledge-aware graph networks for commonsense reasoning}.
\newblock In \emph{Proceedings of the 2019 Conference on Empirical Methods in
  Natural Language Processing and the 9th International Joint Conference on
  Natural Language Processing (EMNLP-IJCNLP)}, pages 2829--2839, Hong Kong,
  China. Association for Computational Linguistics.

\bibitem[{Ma et~al.(2019)Ma, Francis, Lu, Nyberg, and
  Oltramari}]{ma-etal-2019-towards}
Kaixin Ma, Jonathan Francis, Quanyang Lu, Eric Nyberg, and Alessandro
  Oltramari. 2019.
\newblock \href {https://doi.org/10.18653/v1/D19-6003} {Towards generalizable
  neuro-symbolic systems for commonsense question answering}.
\newblock In \emph{Proceedings of the First Workshop on Commonsense Inference
  in Natural Language Processing}, pages 22--32, Hong Kong, China. Association
  for Computational Linguistics.

\bibitem[{Murugesan et~al.(2021{\natexlab{a}})Murugesan, Atzeni, Kapanipathi,
  Shukla, Kumaravel, Tesauro, Talamadupula, Sachan, and Campbell}]{twc}
Keerthiram Murugesan, Mattia Atzeni, Pavan Kapanipathi, Pushkar Shukla, Sadhana
  Kumaravel, Gerald Tesauro, Kartik Talamadupula, Mrinmaya Sachan, and Murray
  Campbell. 2021{\natexlab{a}}.
\newblock {Text-based RL Agents with Commonsense Knowledge: New Challenges,
  Environments and Baselines}.
\newblock In \emph{Thirty Fifth AAAI Conference on Artificial Intelligence}.

\bibitem[{Murugesan et~al.(2021{\natexlab{b}})Murugesan, Atzeni, Kapanipathi,
  Talamadupula, Sachan, and Campbell}]{efficient}
Keerthiram Murugesan, Mattia Atzeni, Pavan Kapanipathi, Kartik Talamadupula,
  Mrinmaya Sachan, and Murray Campbell. 2021{\natexlab{b}}.
\newblock \href {https://doi.org/10.18653/v1/2021.acl-short.91} {Efficient
  text-based reinforcement learning by jointly leveraging state and commonsense
  graph representations}.
\newblock In \emph{Proceedings of the 59th Annual Meeting of the Association
  for Computational Linguistics and the 11th International Joint Conference on
  Natural Language Processing (Volume 2: Short Papers)}, pages 719--725,
  Online. Association for Computational Linguistics.

\bibitem[{Nair and Hinton(2010)}]{relu}
Vinod Nair and Geoffrey~E Hinton. 2010.
\newblock Rectified linear units improve restricted boltzmann machines.
\newblock In \emph{Proceedings of the 27th International Conference on
  International Conference on Machine Learning}, ICML'10, page
  807^^e2^^80^^93814, Madison, WI, USA. Omnipress.

\bibitem[{Paszke et~al.(2017)Paszke, Gross, Chintala, Chanan, Yang, DeVito,
  Lin, Desmaison, Antiga, and Lerer}]{pytorch}
Adam Paszke, Sam Gross, Soumith Chintala, Gregory Chanan, Edward Yang, Zachary
  DeVito, Zeming Lin, Alban Desmaison, Luca Antiga, and Adam Lerer. 2017.
\newblock Automatic differentiation in pytorch.
\newblock In \emph{NIPS-W}.

\bibitem[{Pennington et~al.(2014)Pennington, Socher, and Manning}]{glove}
Jeffrey Pennington, Richard Socher, and Christopher Manning. 2014.
\newblock \href {https://doi.org/10.3115/v1/D14-1162} {{G}lo{V}e: Global
  vectors for word representation}.
\newblock In \emph{Proceedings of the 2014 Conference on Empirical Methods in
  Natural Language Processing ({EMNLP})}, pages 1532--1543, Doha, Qatar.
  Association for Computational Linguistics.

\bibitem[{Qasemi et~al.(2022)Qasemi, LIU, Chen, and Szekely}]{qasemi2022paco}
Ehsan Qasemi, F~LIU, Muhao Chen, and Pedro Szekely. 2022.
\newblock Paco: Preconditions attributed to commonsense knowledge.
\newblock EMNLP.

\bibitem[{Rudinger et~al.(2020)Rudinger, Shwartz, Hwang, Bhagavatula, Forbes,
  Le~Bras, Smith, and Choi}]{rudinger-etal-2020-thinking}
Rachel Rudinger, Vered Shwartz, Jena~D. Hwang, Chandra Bhagavatula, Maxwell
  Forbes, Ronan Le~Bras, Noah~A. Smith, and Yejin Choi. 2020.
\newblock \href {https://doi.org/10.18653/v1/2020.findings-emnlp.418} {Thinking
  like a skeptic: Defeasible inference in natural language}.
\newblock In \emph{Findings of the Association for Computational Linguistics:
  EMNLP 2020}, pages 4661--4675, Online. Association for Computational
  Linguistics.

\bibitem[{Sahith et~al.(2020)Sahith, Spencer, Prithviraj, and Riedl}]{wordplay}
Dambekodi Sahith, Frazier Spencer, Ammanabrolu Prithviraj, and Mark~O. Riedl.
  2020.
\newblock Playing text-based games with common sense.
\newblock In \emph{Advances in Neural Information Processing Systems
  Workshops}.

\bibitem[{Shridhar et~al.(2021)Shridhar, Yuan, C\^ot\'e, Bisk, Trischler, and
  Hausknecht}]{ALFWorld20}
Mohit Shridhar, Xingdi Yuan, Marc-Alexandre C\^ot\'e, Yonatan Bisk, Adam
  Trischler, and Matthew Hausknecht. 2021.
\newblock \href {https://arxiv.org/abs/2010.03768} {{ALFWorld: Aligning Text
  and Embodied Environments for Interactive Learning}}.
\newblock In \emph{Proceedings of the International Conference on Learning
  Representations (ICLR)}.

\bibitem[{Speer et~al.(2017)Speer, Chin, and Havasi}]{conceptnet}
Robyn Speer, Joshua Chin, and Catherine Havasi. 2017.
\newblock Conceptnet 5.5: An open multilingual graph of general knowledge.
\newblock In \emph{Proceedings of the Thirty-First AAAI Conference on
  Artificial Intelligence}, AAAI'17, page 4444^^e2^^80^^934451. AAAI Press.

\bibitem[{Tanaka et~al.(2022)Tanaka, Kimura, and Tatsubori}]{aaaiw}
Tsunehiko Tanaka, Daiki Kimura, and Michiaki Tatsubori. 2022.
\newblock {Commonsense Knowledge from Scene Graphs for Textual Environments}.
\newblock In \emph{Thirty Fifth AAAI Conference on Artificial Intelligence
  Workshop (AAAIW)}.

\bibitem[{Wang et~al.(2022)Wang, Jansen, C{\^o}t{\'e}, and
  Ammanabrolu}]{scienceworld2022}
Ruoyao Wang, Peter Jansen, Marc-Alexandre C{\^o}t{\'e}, and Prithviraj
  Ammanabrolu. 2022.
\newblock \href {http://arxiv.org/abs/2203.07540} {Scienceworld: Is your agent
  smarter than a 5th grader?}

\bibitem[{Zhou et~al.(2021)Zhou, Naseem, Fernandez~Astudillo, Lee, Florian, and
  Roukos}]{amr}
Jiawei Zhou, Tahira Naseem, Ram{\'o}n Fernandez~Astudillo, Young-Suk Lee, Radu
  Florian, and Salim Roukos. 2021.
\newblock \href {https://doi.org/10.18653/v1/2021.emnlp-main.507}
  {Structure-aware fine-tuning of sequence-to-sequence transformers for
  transition-based {AMR} parsing}.
\newblock In \emph{Proceedings of the 2021 Conference on Empirical Methods in
  Natural Language Processing}, pages 6279--6290, Online and Punta Cana,
  Dominican Republic. Association for Computational Linguistics.

\end{thebibliography}
\bibliographystyle{acl_natbib}

\clearpage
\appendix
\section{Appendix}
\subsection{Additional Results}
\label{apx:additionalresults}
\noindent \textbf{Hidden Size}:
We investigate the optimal hidden size for \method~by training five agents with different hidden sizes at the medium level and testing them in the \textit{OUT} set. The results are shown in Fig.~\ref{tab:hidden}. Since we could not find a consistent trend, we used 512 in this work, which had the best performance throughout the experiments in this work.
\begin{table}[h]
  \small
  \centering
  \caption{Ablation results to evaluate the hidden size of \method. All results are in the medium difficulty level using the \textit{OUT} set.}
  \begin{tabular}[t]{>{\centering}m{0.25\linewidth}>{\centering\arraybackslash}m{0.25\linewidth}}
    \toprule
    Hidden size & Scores \\
    \midrule
    128 & 0.71 $\pm$ 0.02 \\
    256 & 0.70 $\pm$ 0.02 \\
    300 & 0.67 $\pm$ 0.02 \\
    512 & \textbf{0.76 $\pm$ 0.02} \\
    1024 & 0.68 $\pm$ 0.03 \\
    \bottomrule
  \end{tabular}
  \label{tab:hidden}
\end{table}

\noindent \textbf{Activation Function in Difference Graph Encoder}: 
We investigate the effectiveness of the activation function in our proposed difference graph encoder (Section~\ref{sec:dge}) added from the base GIN~\cite{gin}. Table~\ref{tab:activationfunction} shows that the performance is better with the activation function than without.
\begin{table}[h]
  \small
  \centering
  \caption{Ablation results to evaluate the activation function in the difference graph encoder (Section~\ref{sec:dge}). All results are in the medium difficulty using the \textit{OUT} set.}
  \begin{tabular}[t]{>{\raggedright}m{0.45\linewidth}>{\centering\arraybackslash}m{0.2\linewidth}}
    \toprule
    Method & Scores \\
    \midrule
    \method~(w/o activation) & 0.64 $\pm$ 0.01 \\
    \method & \textbf{0.76 $\pm$ 0.02} \\
    \bottomrule
  \end{tabular}
  \label{tab:activationfunction}
\end{table}

\newpage
\noindent \textbf{Common Sense Source}: 
We investigate common sense sources. Table~\ref{tab:source} shows that in the hard difficulty level using the \textit{OUT} set, TWC~agent with ConceptNet performs better than that with VG, but \method~does the opposite. \citet{aaaiw} propose that Visual Genome contains more grounded common sense, but TWC~agent shows that it does not have the structure to take advantage of it. In contrast, \method~is able to fully utilize VG. Based on this result, we use VG in our experiments.
\begin{table}[h]
  \small
  \centering
  \caption{Ablation results to evaluate common sense sources. All results are in the hard difficulty using the \textit{OUT} set. CN and VG denote ConceptNet~\cite{conceptnet} and Visual Genome~\cite{visualgenome}, respectively.}
  \begin{tabular}[t]{>{\raggedright}m{0.3\linewidth}>{\centering}m{0.1\linewidth}>{\centering\arraybackslash}m{0.2\linewidth}}
    \toprule
    Method & Source & Scores \\
    \midrule
    TWC~agent-CN & CN & 0.29 $\pm$ 0.02 \\
    TWC~agent-VG & VG & 0.25 $\pm$ 0.02 \\
    \method-CN & CN & 0.31 $\pm$ 0.02 \\
    \method-VG & VG & \textbf{0.35 $\pm$ 0.02} \\
    \bottomrule
  \end{tabular}
  \label{tab:source}
\end{table}

\end{document}